\documentclass[10pt, a4paper]{article}

\usepackage{lrec-coling2024} % this is the new style
\usepackage{inconsolata}
\usepackage{pgfplots}
\pgfplotsset{compat=1.17}

\title{EmoMix-3L: A Code-Mixed Dataset for Bangla-English-Hindi Emotion Detection}

\name{Nishat Raihan\textsuperscript{*}, Dhiman Goswami\textsuperscript{*}, Antara Mahmud, \\ \textbf{\large Antonios Anastasopoulos, Marcos Zampieri}\\} 
\address{George Mason University, USA \\
         \{mraihan2, dgoswam\}@gmu.edu \\}

\abstract{
Code-mixing is a well-studied linguistic phenomenon that occurs when two or more languages are mixed in text or speech. Several studies have been conducted on building datasets and performing downstream NLP tasks on code-mixed data. Although it is not uncommon to observe code-mixing of three or more languages, most available datasets in this domain contain code-mixed data from only two languages. In this paper, we introduce EmoMix-3L, a novel multi-label emotion detection dataset containing code-mixed data from three different languages. We experiment with several models on EmoMix-3L and we report that MuRIL outperforms other models on this dataset. \\
\newline 
}

\begin{document}

\maketitleabstract

\section{Introduction}

The ability to convey emotions is an essential part of human communication. NLP models have been applied to detect emotions (e.g., anger, fear, joy) in texts from social media \cite{gaind2019emotion}, customer service \cite{gupta2010emotion}, and healthcare \cite{ayata2020emotion}. Emotion detection is an important part of social media analysis and mining efforts that include popular tasks such as sentiment analysis \cite{liu2020sentiment} and stance detection \cite{kawintiranon2021knowledge}. 

Most studies on sentiment analysis and emotion detection are carried out in one language at a time \cite{abdul2017emonet,chatterjee2019semeval}. Apart from a few notable exceptions \cite{vedula2023precogiiith}, detecting emotion in multilingual and code-mixed environments has not been significantly explored. Code-mixing is very common in multilingual societies. It is defined as the practice of using words and grammatical constructions from two or more languages interchangeably \cite{muysken2000study}. Code-mixing can occur at various levels such as intra-sentential where code-mixing is present within a sentence, and inter-sententialwhere code-mixing is present across sentences.

Detecting sentiments and emotions in code-mixed texts is a challenging task that we address in this paper by introducing EmoMix-3L, a multi-label emotion detection containing Bangla, Hindi, and English code-mixed texts. These three languages are often used together by the population of West Bengal. They are also used by populations from South East Asian living in other parts of the world where English is spoken as the official language or \emph{lingua franca} such London, New York, or Singapore. Recent studies have created resources for these three languages in tasks such as sentiment analysis and offensive language detection \cite{nishat2023sentmix,goswami2023offmix}. To the best of our knowledge, however, no datasets for emotion detection in Bangla-English-Hindi code-mix exists and EmoMix-3L fills this gap. 

The main contributions of this paper are as follows:

\begin{itemize}
    \item We introduce EmoMix-3L\footnote{\url{https://github.com/GoswamiDhiman/EmoMix-3L}}, a novel three-language code-mixed test dataset in Bangla-Hindi-English for multi-label emotion detection. EmoMix-3L contains 1,071 instances annotated by speakers of the three languages. We make EmoMix-3L freely available to the community.
    \item We provide a comprehensive evaluation of several monolingual, bilingual, and multilingual models on EmoMix-3L.
\end{itemize}

\noindent We present EmoMix-3L exclusively as a test set due to the unique and specialized nature of the task. The size of the dataset, while limited for training purposes, offers a high-quality testing benchmark with gold-standard labels. Given the scarcity of similar datasets and the challenges associated with data collection, EmoMix-3L provides an important resource for the evaluation of multi-label emotion detection models, filling a critical gap in multi-level code-mixing research. 

\section{Related Work}

A few studies studies have addressed emotion detection on bilingual code-mixed data \cite{wadhawan2021towards,vedula2023precogiiith,ameer2022multi}. \citet{vedula2023precogiiith} implemented a multi-class emotion detection model leveraging transformer-based multilingual Large Language Models (LLMs) for English-Urdu code-mixed text. However, the study's ability to interpret code-mixed sentences that combine English and Roman Urdu had limitations. The study by \citet{ameer2022multi} highlights how multi-label emotion classification may be used to identify every emotion that could exist in a given text. 11,914 code-mixed (English and Roman Urdu) SMS messages make up the substantial benchmark corpus presented in this paper for the multi-label emotion classification challenge.

% \citet{wadhawan2021towards} used a CNN-BiLSTM model and a pre-trained bilingual model to recognize emotions in Hindi and English code-mixed. The authors introduced a class-balanced dataset of code-mixed Hindi-English data for emotion identification. The study demonstrates the application of transformer-based models and deep learning for emotion detection in tweets with mixed Hindi-English codes. The transformer-based BERT model performs the best, surpassing all other models.

% \citet{wadhawan2021towards} introduced a deep learning strategy that uses deep neural networks, specifically the CNN-BiLSTM model, and a pre-trained bilingual model to recognize emotions conveyed with Hindi-English code-mixed social media posts. An openly available class-balanced dataset of code-mixed Hindi-English data for emotion identification is provided by \citet{wadhawan2021towards}. The study also demonstrates the application of transformer-based models and deep learning for emotion detection in tweets with mixed Hindi-English codes. The transformer-based BERT model performs the best, surpassing all other models.

There have been a number studies on Bengali-English code-mixed data. \citet{mursalin2022deep} used deep learning approaches to identifying emotions from texts containing mixed Bengali and English coding, with an emphasis on comparing and contrasting the effectiveness of the suggested model with other ML and DL methods already in use. \citet{ahmad2019review} have explored and analyzed regional Indian code-mixed data. In this paper, the importance and applications of sentiment detection in a variety of domains are discussed, with an emphasis on Indo-Aryan languages like Tamil, Bengali, and Hindi.

A few studies have addressing Bengali-English-Hindi code-mixing on social media. \citet{nishat2023sentmix} uses multiple monolingual, bilingual, and multilingual models and a unique dataset with gold standard labels for sentiment analysis in Bangla-English-Hindi. \citet{goswami2023offmix} presents a novel offensive language identification dataset with the same three languages. Finally, another similar work include the TB-OLID dataset \cite{raihan2023offensive} that contains both transliterated and code-mixed data for offensive language identification. 

%While datasets with code-mixing of two languages has been substantially explored, to the best of our knowledge, EmoMix-3L is the first emotion detection dataset to include code-mixing between three languages. 

% In the field of NLP research, there is a lack of studies on understanding emotions in bilingual code-mixed data. Code-mixing is when people use two or more languages together in the same conversation. Most research focuses on only two languages, but there is not much work on code-mixing with three or more languages. This work focuses on emotion detection in conversations that mix three languages. By going beyond just two languages, we hope to uncover the complicated patterns in how people use multiple languages to express their feelings. 

\section{The \textit{EmoMix-3L} Dataset}

We choose a controlled data collection method, asking the volunteers to freely contribute data in Bangla, English, and Hindi. This decision stems from several challenges of extracting such specific code-mixed data from social media and other online platforms. Our approach ensures data quality and sidesteps the ethical concerns associated with using publicly available online data. Such types of datasets are often used when it is difficult to mine them from existing corpora. As examples, for fine-tuning LLMs on instructions and conversations, semi-natural datasets like \citet{dolly2023} and \citet{awesome_instruction_datasets} have become popular. 

\paragraph{Data Collection} 

Ten undergraduate students fluent in the three languages was asked to prepare 300 to 350 social media posts each. They were allowed to use any language, including Bangla, English, and Hindi to prepare posts on several daily topics like politics, sports, education, social media rumors, etc. We also ask them to switch languages if and wherever they feel comfortable doing so. The inclusion of emojis, hashtags, and transliteration was also encouraged. The students had the flexibility to prepare the data as naturally as possible. Upon completion of this stage, we gathered 2,598 samples that contained at least one word or sub-word from each of the three languages using langdetect \cite{langdetect} an open-sourced Python tool for language identification.  

\paragraph{Data Annotation} We annotate the dataset in two steps. Firstly, we recruited three students from social science, computer science, and linguistics fluent in the three languages to serve as annotators. They annotated all 2,598 samples with one of the five labels (Happy, Surprise, Neutral, Sad, Angry) with a raw agreement of 47.9\%. We then take 1,246 instances, where all three annotators agree on the labels, and use them in a second step. To further ensure high-quality annotation, we recruit a second group of annotators consisting of two NLP researchers fluent in the three languages. After their annotation, we calculate a raw agreement of 86\% \cite{kvaalseth1989note}, a Cohen Kappa score of 0.72. After the two stages, we only keep the instances where both annotators agree, and we end up with a total of 1,071 instances. The label distribution is shown in Table \ref{tab:label_distribution}. 

\begin{table} [!h]
\centering
\scalebox{.92}{
\begin{tabular}{lcc}
\hline
\textbf{Label} & \textbf{Instances} & \textbf{Percentage}\\
\hline
Happy & 228 & 21.29\% \\
Surprise & 227 & 21.20\% \\
Neutral & 223 & 20.82\% \\
Sad & 205 & 19.14\% \\
Angry & 188 & 17.55\% \\

\hline
Total & 1,071 & 100\% \\
\hline
\end{tabular}
}
\caption{Label distribution in EmoMix-3L}
\label{tab:label_distribution}
\end{table}

\paragraph{Dataset Statistics} A detailed description of the dataset statistics is provided in Table \ref{tab:data_card}. Since the dataset was generated by people whose first language is Bangla, we observe that the majority of tokens in the dataset are in Bangla. There are several \textit{Other} tokens in the dataset that are not from Bangla, English, or Hindi language. The \textit{Other} tokens in the dataset primarily contain transliterated words as well as emojis and hashtags. Also, there are several misspelled words that have been classified as \textit{Other} tokens too. 
%We have Positive and Negative labels for the 1007 data. 

\begin{table} [!h]
\centering
\scalebox{.82}{
\begin{tabular}{l|c|cccc}
\hline
 & \textbf{All} & \textbf{Bangla} & \textbf{English} & \textbf{Hindi} & \textbf{Other}\\
\hline
Tokens & 98,011 & 36,784 & 6,587 & 15,560 & 39,080\\
Types & 21,766 & 9,118 & 1,237 & 1,523 & 10,022\\
% Max. in instance & 173 & 62 & 20 & 47  & 93\\
% Min. in instance & 41 & 4 & 3 & 2 & 8\\
Avg & 91.51 & 34.35 & 6.15 & 14.53 & 36.49\\
Std Dev & 20.24 & 9.13 & 2.88 & 5.94 & 10.64\\
\hline
\end{tabular}
}
\caption{EmoMix-3L Data Card. The row {\em Avg} represents the average number of tokens with its standard deviation in row {\em Std Dev}.}
\label{tab:data_card}
\end{table}

\paragraph{Synthetic Train and Development Set} We present EmoMix-3L as a test dataset and we build a synthetic train and development set that contains Code-mixing for Bangla, English, and Hindi. We use an English training dataset annotated with the same labels as EmoMix-3L, namely Social Media Emotion Dataset (SMED)\footnote{\url{https://www.kaggle.com/datasets/gangulyamrita/social-media-emotion-dataset}}. We then use the \textit{Random Code-mixing Algorithm} \cite{krishnan2022cross} and \textit{r-CM}  \cite{santy2021bertologicomix} to generate the synthetic Code-mixed dataset. Similar approach is also found in \cite{gautam2021comet}.
%Finally, we split the dataset using a 60-20-20 split with balanced label distribution to serve as training, development, and test sets. 
%Finally, we split the dataset in an 80-20 way with balanced label distribution to prepare the train and development dataset. 
%(Appendix \ref{Appendix_B}).

\section{Experiments}

\paragraph{Monolingual Models} We use six monolingual models for these experiments, five general models, and one task fined-tuned model. The five monolingual models are DistilBERT \cite{DBLP:journals/corr/abs-1910-01108}, BERT \cite{devlin2019bert}, BanglaBERT \cite{kowsher2022bangla}, roBERTa \cite{liu2019roberta}, HindiBERT \cite{nick_doiron_2023}. BanglaBERT is trained in only Bangla and HindiBERT in only Hindi while DistilBERT, BERT, and roBERTa are trained in English only. Finally, the English task fine-tuned model we use is emoBERTa \cite{kim2021emoberta}.

\paragraph{Bilingual Models} BanglishBERT \cite{bhattacharjee-etal-2022-banglabert} and HingBERT \cite{nayak-joshi-2022-l3cube} are used as bilingual models as they are trained on both Bangla-English and Hindi-English respectively.

\paragraph{Multilingual Models} We use mBERT \cite{devlin2019bert} and XLM-R \cite{conneau2019unsupervised} as multilingual models which are respectively trained on 104 and 100 languages including Bangla-English-Hindi. We also use IndicBERT \cite{kakwani2020indicnlpsuite} and MuRIL \cite{khanuja2021muril} which cover 12 and 17 Indian languages, respectively, including Bangla-English-Hindi. We also perform hyper-parameter tuning while using all the models to prevent overfitting.

\paragraph{Prompting} We use prompting with GPT-3.5-turbo model \cite{openai2023gpt35turbo} from OpenAI for this task. We use the API for zero-shot prompting (see Figure \ref{fig:prompt1}) and ask the model to label the test set.\\

Additionally, we run the same experiments separately on synthetic and natural datasets splitting both in a 60-20-20 way for training, evaluating, and testing purposes.

\begin{figure}[h]
\centering
\scalebox{.92}{
\begin{tikzpicture}[node distance=1cm]
    % Styles for nodes
    \tikzstyle{block} = [rectangle, draw, fill=cyan!20, text width=\linewidth, text centered, rounded corners, minimum height=4em]
    \tikzstyle{operation} = [text centered, minimum height=1em]
    % Nodes
    \node [block] (rect1) {\textbf{Role:}{ "You are a helpful AI assistant. You are given the task of Emotion Detection. }};
    \node [operation, below of=rect1] (plus1) {};
    \node [block, below of=plus1] (rect2) {\textbf{Definition:}{ An emotion is a feeling that can be caused by the situation that people are in or the people they are with. You will be given a text to label either 'Happy', 'Surprise', 'Neutral', 'Sad' or 'Angry'. }};
    \node [operation, below of=rect2] (plus2) {};
    \node [block, below of=plus2] (rect3) {\textbf{Task:}{ Generate the label for this \textbf{"text"} in the following format: \textit{<label> Your\_Predicted\_Label <$\backslash$label>}. Thanks."}};
\end{tikzpicture}
}
\caption{Sample GPT-3.5 prompt.}
\label{fig:prompt1}
\end{figure}

\section{Results}

In this experiment, synthetic data is used as a training set, and natural data is used as the test set. The F1 scores of monolingual models range from 0.14 to 0.41, where roBERTa performs the best. MuRIL is the best of all the multilingual models, with an F1 score of 0.54. Besides, a zero-shot prompting technique on GPT 3.5 turbo provides a 0.51 weighted F1 score. The task fine-tuned model emoBERTa provides the F1 score of 0.42. BanglishBERT scores 0.44 which is the best F1 score among all the bilingual models. These results are available in Table \ref{Results1}. 
%and Appendix \ref{Appendix_A}.

\begin{table} [!h]
\centering
\scalebox{0.92}{
\begin{tabular}{lcc}
\hline
\textbf{Models}  & \textbf{F1 Score} \\
\hline
MuRIL & \textbf{0.54} \\
XLM-R & 0.51 \\
GPT 3.5 Turbo & 0.51 \\
BanglishBERT & 0.44 \\
HingBERT & 0.43  \\
emoBERTa & 0.42\\
roBERTa & 0.41  \\
BERT & 0.38 \\
mBERT & 0.35 \\
DistilBERT & 0.24 \\
IndicBERT & 0.22 \\
BanglaBERT & 0.16 \\
HindiBERT & 0.14  \\
\hline
\end{tabular}
}
\caption{Weighted F-1 score for different models: training on synthetic and tested on natural data (EmoMix-3L).}
\label{Results1}
\end{table}

\noindent We perform the same experiment using synthetic data for training and testing. We present results in Table \ref{tab_synthetic}. Here, MuRIL with 0.67 F1 score is the best-performing model. BERT is the best among the monolingual models where their F1 range from 0.32 to 0.45. BanglishBERT with 0.47 F1 score is the best among the bilingual models. The task fine-tuned model emoBERTa scores 0.41 for the synthetic dataset.

\begin{table} [!h]
\centering
\scalebox{.92}{
\begin{tabular}{lcc}
\hline
\textbf{Models}  & \textbf{Weighted F1 Score} \\
\hline
MuRIL & \textbf{0.67} \\
XLM-R & 0.51 \\
mBERT & 0.49 \\
BanglishBERT & 0.47 \\
HingBERT & 0.45  \\
BERT & 0.44 \\
emoBERTa & 0.41 \\
roBERTa & 0.41  \\
DistilBERT & 0.40  \\
BanglaBERT & 0.39 \\
IndicBERT & 0.38 \\
HindiBERT & 0.32  \\
\hline
\end{tabular}
}
\caption{Weighted F-1 score for different models: training on synthetic and tested on synthetic data.}
\label{tab_synthetic}
\end{table}

\subsection{Error Analysis}

The confusion matrix for the best-performing model MuRIL for training on synthetic and tested in EmoMix-3L is shown in Figure \ref{fig:cf}.

\begin{figure}[!ht]
  \centering
  \includegraphics[width=\linewidth]{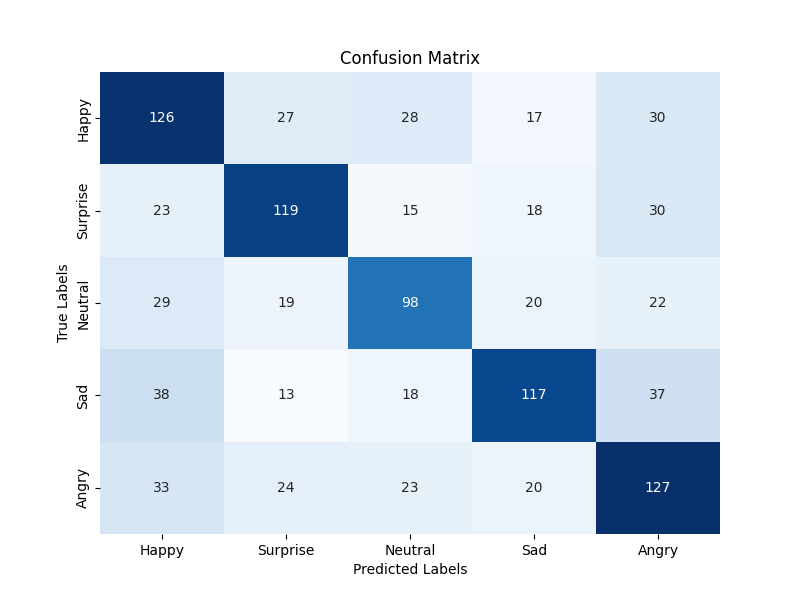}
  \caption{Confusion Matrix (Training on synthetic data, tested on EmoMix-3L).}
  \label{fig:cf}
\end{figure}

\noindent We observe \textit{Other} tokens in more than 39\% of the whole dataset, as shown in Table \ref{tab:data_card}. These tokens occur due to transliteration which poses a challenge for most of the models since not all of the models are pre-trained on transliterated tokens. BanglishBERT did well since it recognizes both Bangla and English tokens. However, the total number of tokens for Hindi-English is less than Bangla-English tokens, justifying HingBERT's inferior performance compared to BanglishBERT (see Table \ref{Results1}). Also, misspelled words and typos are also observed in the datasets, which are, for the most part, unknown tokens for the models, making the task even more difficult. Some examples are available in Appendix \ref{sec:appendix_a} which are classified wrongly by all the models.

\section{Conclusion and Future Work}

We introduce EmoMix-3L, a novel dataset containing 1,071 instances of Bangla-English-Hindi code-mixed content. We have also 
%come up 
created 100,000 instances of synthetic data in the same languages to facilitate our training methods. We have tested 
%a bunch 
multiple monolingual models on these datasets, and MuRIL performs the best, especially when it was trained on synthetic data and tested on EmoMix-3L. MuRIL was also the best in the scenario of both training and testing on synthetic data, outperforming all the other models in multi-label emotion detection. Looking ahead, we would like turning EmoMix-3L into a larger dataset serving both as a training and testing dataset. We would also like to create pre-trained tri-lingual code-mixing models. It will facilitate the emotion detection task in the intricate mix of Bangla, English, and Hindi. Moreover, we would like to explore the performance of large language models by fine-tuning them on code-mixed datasets. This will provide valuable insights into their unexplored training corpora and their ability to cope with code-mixed scenarios.

\section*{References}
\bibliography{custom}
\bibliographystyle{lrec-coling2024-natbib}

\appendix

\section{Examples of Misclassified Instances}
\label{sec:appendix_a}

\begin{figure}[!ht]
  \centering
  \includegraphics[width=0.915\linewidth]{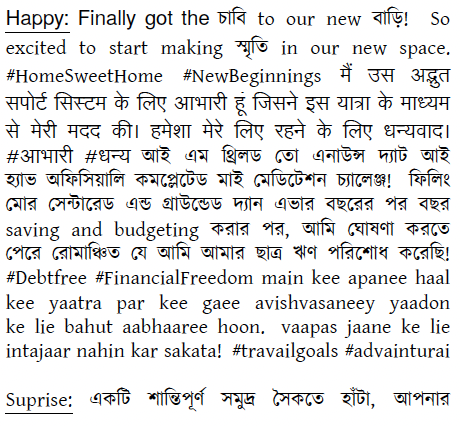}
  %\caption{Confusion Matrix (Training on synthetic data, tested on EmoMix-3L).}
  \label{fig:cf}
\end{figure}

\begin{figure}[!ht]
  \centering
  \includegraphics[width=\linewidth]{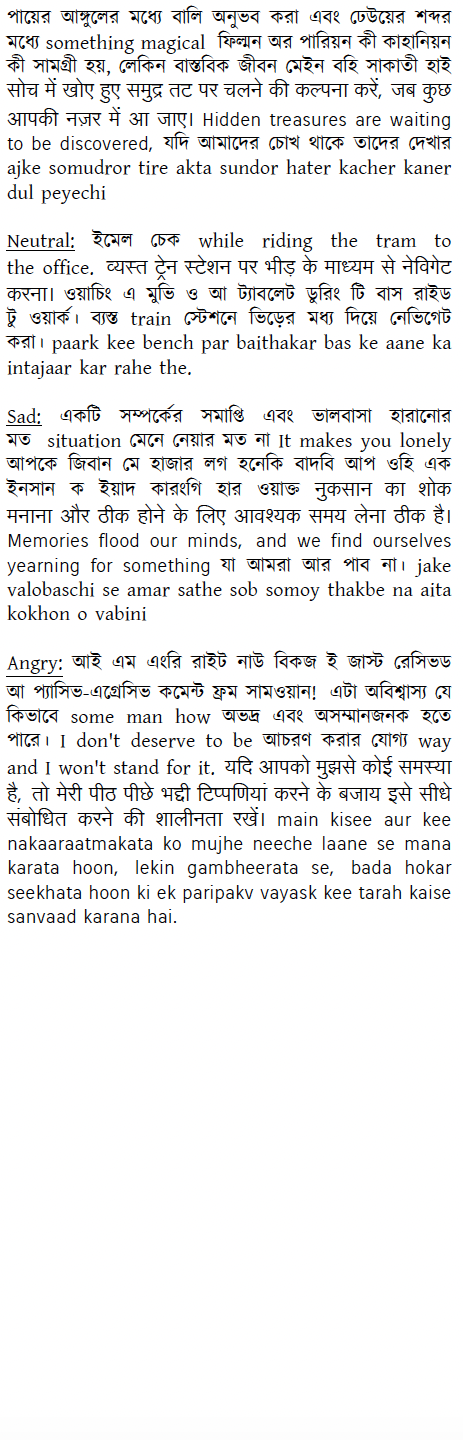}
  %\caption{Confusion Matrix (Training on synthetic data, tested on EmoMix-3L).}
  \label{fig:cf}
\end{figure}

\end{document}